\documentclass[sigconf]{acmart}

\usepackage{graphicx}
\usepackage{url}
\usepackage{booktabs}
\usepackage{array}
\usepackage{xcolor}
\usepackage{colortbl}
\usepackage{threeparttable}
\usepackage{tcolorbox}
\AtBeginDocument{%
  }

\setcopyright{acmlicensed}
\copyrightyear{2025}
\acmYear{2025}
\acmDOI{XXXXXXX.XXXXXXX}
\acmConference[CODS 2025]{International Conference on Data Science}{December 17--20,
  2025}{Pune, India}
\acmISBN{978-1-4503-XXXX-X/2018/06}

\acmSubmissionID{260}

\begin{document}

\title{ExaCraft: Dynamic Learning Context Adaptation for Personalized Educational Examples}


\author{Akaash Chatterjee}
\affiliation{%
  \institution{Indian Institute of Technology Jodhpur}
  \city{Jodhpur}
  \state{Rajasthan}
  \country{India}}
\email{m24cse002@iitj.ac.in}

\author{Suman Kundu}
\affiliation{%
  \institution{Indian Institute of Technology Jodhpur}
  \city{Jodhpur}
  \state{Rajasthan}
  \country{India}}
  \email{suman@iitj.ac.in}

\renewcommand{\shortauthors}{Akaash Chatterjee and Suman Kundu}

\begin{abstract}
  Learning is most effective when it's connected to relevant, relatable examples that resonate with learners on a personal level. However, existing educational AI tools don't focus on generating examples or adapting to learners' changing understanding, struggles, or growing skills. We've developed ExaCraft, an AI system that generates personalized examples by adapting to the learner's dynamic context. Through the Google Gemini AI and Python Flask API, accessible via a Chrome extension, ExaCraft combines user-defined profiles (including location, education, profession, and complexity preferences) with real-time analysis of learner behavior. This ensures examples are both culturally relevant and tailored to individual learning needs. The system's core innovation is its ability to adapt to five key aspects of the learning context: indicators of struggle, mastery patterns, topic progression history, session boundaries, and learning progression signals. Our demonstration will show how ExaCraft's examples evolve from basic concepts to advanced technical implementations, responding to topic repetition, regeneration requests, and topic progression patterns in different use cases.
\end{abstract}



\received{1st September 2025}
\received[accepted]{8th October 2025}

\maketitle

\section{Introduction}

Traditional education often uses generic examples. These do not connect well with learners' different backgrounds, experiences, and ways of thinking. This one-size-fits-all method for generating examples limits learning, as learners often struggle to connect with these examples. Consider teaching machine learning: a standard textbook might explain algorithms using abstract mathematical notation or generic datasets like `iris classification.' However, a marketing professional in Mumbai would better understand these concepts through examples of customer segmentation for e-commerce campaigns, while a biology student in Delhi might grasp the same algorithms more effectively through examples of genetic sequence analysis or disease prediction models. Moreover, the same learner's example preferences evolve throughout their learning journey. 

The critical flaw in today's AI-assisted learning tools is their failure to personalize examples in real-time. While platforms like ALEKS or Duolingo excel at personalizing the sequence of learning tasks, they still present every user with the same static, one-size-fits-all examples from a fixed library. This approach fails to solve the core problem: adapting the teaching content itself to a learner's real-time needs. By not tailoring the examples, these systems can't effectively respond to a user's unique struggles or growing mastery, creating a significant barrier to true comprehension.

\paragraph{Our System}
ExaCraft is an AI system that creates personalized examples for learning, automatically adjusting them to the learners' interaction patterns in real-time. It is a Chrome browser extension that generates contextually-aware, personalized examples for any highlighted text, seamlessly integrating with users' natural web browsing workflows. The system uses a Python Flask API that combines Google Gemini AI with learning analytics to deliver culturally relevant and pedagogically adaptive examples.
The system implements a hybrid personalization approach that combines user-configured static profiles with dynamic behavioral adaptation. The static foundation ensures cultural relevance and professional contextualization, while the dynamic components enable real-time adaptation based on learning behaviors.

\paragraph{Contributions}
This work makes the following key contributions to educational AI research and practice:

\begin{itemize}
    \item \textbf{A hybrid personalization framework} that combines user-configured static profiles with dynamic behavioral adaptation to generate personally relevant and pedagogically appropriate examples.
    
    \item \textbf{A seamless workflow integration} via a browser extension that provides zero-disruption example generation with integrated profile management, addressing critical adoption barriers in educational technology.
    
    
    \item \textbf{A system for cross-session personalization continuity} that maintains both static preferences and dynamic adaptation patterns across multiple browsing sessions, enabling long-term, personalized learning progressions.
    
    \item \textbf{A fully deployable source code.}
\end{itemize}

\begin{table}[h]
\centering
\footnotesize
\caption{Static and Dynamic Personalization Features}
\label{tab:features}
\begin{tabular}{>{\columncolor{blue!8}}p{0.35\columnwidth} >{\columncolor{orange!8}}p{0.45\columnwidth}}
\toprule
\rowcolor{gray!25}
\textbf{Static Profile Features} & \textbf{Dynamic Learning Context} \\
\midrule
\rowcolor{blue!3}
Name & \cellcolor{orange!3} Recent topic requests with timestamps \\
\rowcolor{blue!3}
Location (city, country) & \cellcolor{orange!3} Struggle indicators (topic repetition) \\
\rowcolor{blue!3}
Education level & \cellcolor{orange!3} Mastery signals (rapid progression) \\
\rowcolor{blue!3}
Profession & \cellcolor{orange!3} Regeneration button clicks \\
\rowcolor{blue!3}
Preferred complexity & \cellcolor{orange!3} Session boundary tracking \\
\rowcolor{blue!3}
& \cellcolor{orange!3} Cross-session continuity \\
\rowcolor{blue!3}
& \cellcolor{orange!3} Behavioral signal integration \\
\bottomrule
\end{tabular}
\end{table}

\section{Related Work}

Educational personalization has been extensively studied in adaptive learning systems and intelligent tutoring systems (ITS). Traditional approaches focus on creating detailed learner models based on knowledge states and skill assessments, often using techniques like Bayesian Knowledge Tracing to track mastery over time \cite{10.1007/s11257-017-9193-2}. The effectiveness of these systems compared to human tutoring has been a major area of research, highlighting the importance of adaptive support \cite{VanLEHN01102011}. However, these systems typically rely on explicit assessments within a structured environment rather than implicit behavioral signals during natural learning activities.

Recent work in AI-driven educational content generation has explored using large language models for personalized content creation \cite{KASNECI2023102274}. While powerful, most of these systems primarily use static user profiles to guide generation and do not typically incorporate real-time learning behavior adaptation. Learning analytics research has investigated behavioral pattern recognition in online educational environments to understand student engagement and success \cite{Gasevic}. However, these approaches typically analyze learning patterns post-hoc for course improvement rather than using them for real-time content adaptation.

Struggle detection focuses on predicting whether the learner is facing difficulties in understanding a topic. In educational systems, it has been primarily based on performance metrics such as quiz scores and error rates \cite{Baker}, as well as detecting learners' emotional states through affect-sensitive models \cite{dmello}. Our work extends this by focusing on purely behavioral signals (interaction patterns, regeneration requests) as implicit indicators of learning difficulties, which are then used to adapt content immediately.

\begin{figure*}[t]
  \centering
  \scalebox{0.9}{ 
  \begin{minipage}[t]{0.69\textwidth}
    \centering
    \includegraphics[width=\textwidth]{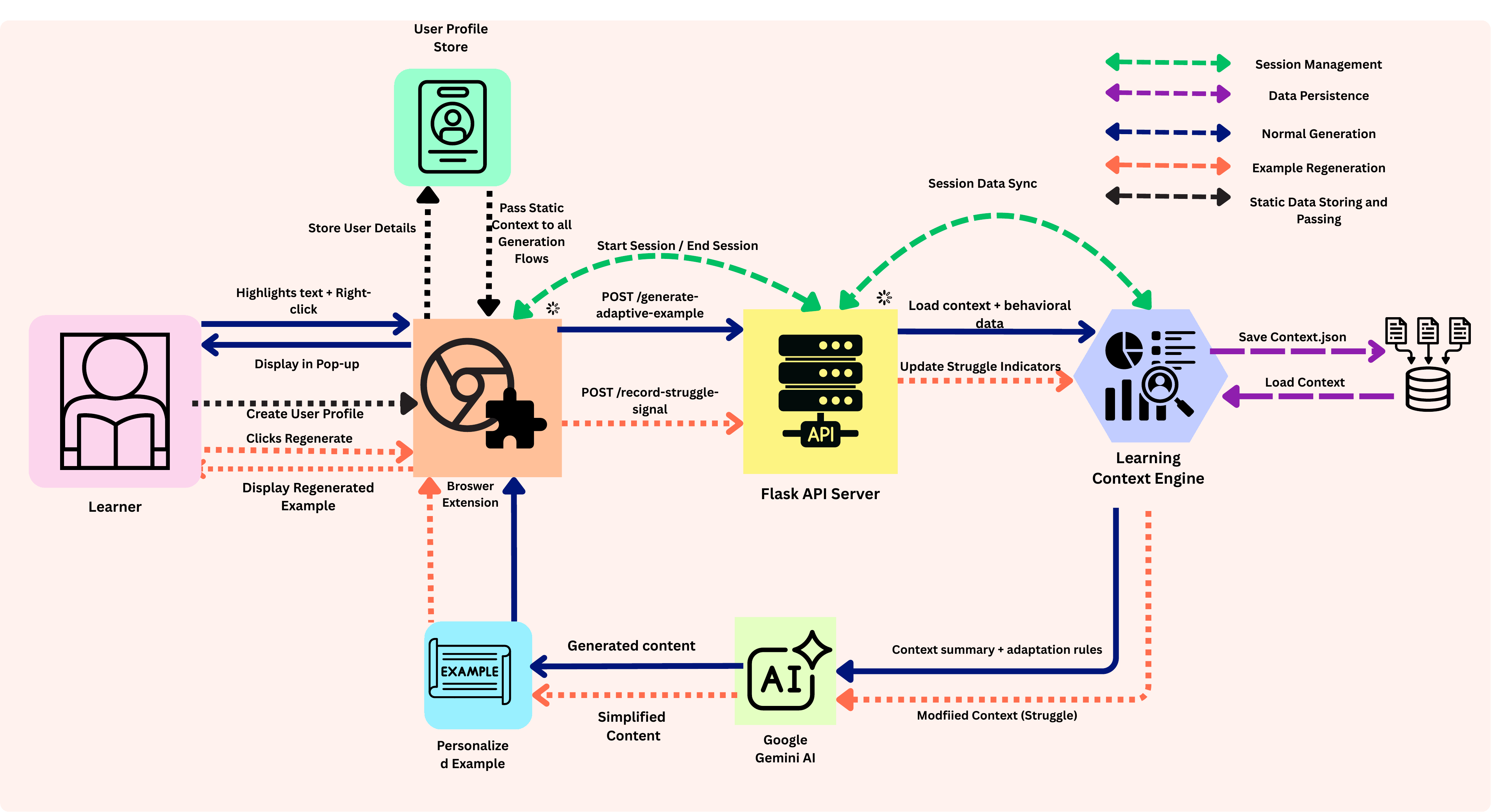}
  \end{minipage}\hfill
  \begin{minipage}[t]{0.30\textwidth}
    \begin{tcolorbox}[colback=teal!5!white, colframe=teal!75!black,
      title=Core Prompt Template]
      \footnotesize
      \textbf{Input:} User profile, dynamic learning context, target topic. \\[3pt]
      \textbf{Prompt Summary:} The prompt instructs the AI to generate adaptive examples based on both static user profile and dynamic learning context. It includes explicit instructions for adapting to struggle indicators (simplifying content, using concrete analogies), mastery indicators (increasing complexity, making connections), and recent topic connections. The hierarchy prioritizes learning context adaptation, followed by cultural personalization, then professional relevance. \\[3pt]
      \textbf{Output:} A contextually adaptive example as a vivid scenario in 2-4 sentences, with specific characters, locations, and situations matching the user's profile and learning state.
    \end{tcolorbox}
  \end{minipage}
  } 
  \caption{ExaCraft system architecture and core prompt template. 
  The left panel illustrates multi-flow interactions across the browser extension, API server, and learning context engine, while the right panel presents the \textit{Core Prompt Template}.}
  \label{fig:system-architecture}
\end{figure*}

\section{System Architecture and Implementation}
Figure~\ref{fig:system-architecture} illustrates the overall ExaCraft system architecture, which integrates static and dynamic personalization features summarized in Table~\ref{tab:features}. 
The left panel depicts the multi-flow interaction between the browser extension, API server, and learning context engine. It shows four different operational flows: normal example generation (blue arrows), struggle detection and regeneration (orange arrows), session management (green arrows), and data persistence (purple arrows). The right panel shows the \textit{Core Prompt Template} used to construct adaptive prompts for the Google Gemini API. 
This template operationalizes the adaptation logic by combining user profile attributes with dynamic behavioral signals to produce contextually adaptive examples in real time. This section provides the details of all four flows.

\subsection{Core Components}
ExaCraft consists of three integrated components as described here.

\textbf{Browser Extension:} A Chrome extension provides smooth integration with web browsing workflows. Users highlight text and generate contextual examples via right-click menus. The extension features a user profile management interface where users specify the static personalization fields detailed in Table~\ref{tab:features}. Beyond profile management, it tracks real-time behavior signals that enable dynamic adaptation. As shown in Figure \ref{fig:system-architecture}, the extension handles both regular generation requests (blue flow) and regeneration signals (orange flow).

\textbf{API Server:} A Python Flask RESTful backend service orchestrates AI content generation with behavioral analytics. It receives requests from the extension, processes user profile data, tracks learning contexts, and integrates with Google Gemini AI through LangChain for refined prompt engineering. The API server processes these inputs and coordinates with the learning context engine to deliver examples adapted to user behavior.

\textbf{Learning Context Engine:} This novel component tracks and analyses the dynamic learning context categories listed in Table~\ref{tab:features}. Its core technical innovation is continuously analysing interaction patterns to build contextual models for real-time personalization. The engine detects learning struggles through topic repetition and regeneration signals, while identifying mastery through rapid topic progression patterns. It manages session boundaries and continuity (green flow) by tracking when users start new learning sessions and maintaining context across session breaks. This engine maintains persistent storage through JSON files (purple flow) for cross-session continuity.


\begin{table}[h]
\centering
\footnotesize
\caption{Adaptation Logic: Mapping Triggers to Actions}
\label{tab:adaptation_rules}
\begin{tabular}{>{\columncolor{blue!8}}p{0.45\columnwidth} >{\columncolor{orange!8}}p{0.45\columnwidth}}
\toprule
\rowcolor{gray!25}
\textbf{Trigger Condition} & \textbf{Adaptation Action} \\
\midrule
\rowcolor{blue!3}
Topic request repetition $\geq$ 3 times & \cellcolor{orange!3} Reduce example complexity, add foundational analogies \\
\rowcolor{blue!3}
Regeneration requests & \cellcolor{orange!3} Simplify content, use more concrete examples \\
\rowcolor{blue!3}
Rapid topic progression (3+ different topics) & \cellcolor{orange!3} Increase complexity, add advanced applications \\
\rowcolor{blue!3}
Session active with multiple topics & \cellcolor{orange!3} Introduce cross-topic connections \\
\rowcolor{blue!3}
Return to previously struggled topic & \cellcolor{orange!3} Apply previously effective simplification strategies \\
\bottomrule
\end{tabular}
\end{table}




\subsection{Hybrid Static-Dynamic Personalization}
Users configure their static profiles through the browser extension interface, providing the contextual information outlined in Table~\ref{tab:features}. These details enable culturally relevant and professionally appropriate examples, as demonstrated in Tables~\ref{tab:profession} and~\ref{tab:education}.  
Furthermore, the system continuously monitors user interactions to detect learning patterns. The engine captures timestamps for each topic request, counts regeneration attempts per topic, and maintains topic-to-topic progression sequences to identify learning velocity. Session boundaries are managed on the user end through explicit session controls, while cross-session patterns are reconstructed from persistent user interaction histories. When users exhibit struggle indicators, as outlined in Table~\ref{tab:features}, ExaCraft automatically reduces complexity. Conversely, mastery signals trigger increased complexity while maintaining the contextual relevance established through static profiling. These adaptation patterns are systematically described in Table~\ref{tab:adaptation_rules}, which summarizes how each contextual trigger maps to a concrete modification in the generated example

\begin{table}[t]
\centering
\footnotesize
\begin{threeparttable}
\caption{Example Variations Based on Professional Background}
\label{tab:profession}
\begin{tabular}{>{\raggedright\columncolor{gray!8}}p{0.18\columnwidth} >{\raggedright\arraybackslash}p{0.32\columnwidth} >{\raggedright\arraybackslash}p{0.32\columnwidth}}
\toprule
\rowcolor{gray!25}
\textbf{Topic} & \textbf{Raj (Mechanic, Bhilai)} & \textbf{Emma (Marketing Manager, Toronto)} \\
\midrule
\rowcolor{blue!3}
\textbf{Machine Learning} & "Raj, a mechanic in Bhilai, notices trucks from the steel plant break down more during monsoons. He wonders if tracking breakdown patterns and weather data could predict maintenance needs, preventing delays for factory workers." & "Emma, a marketing manager in Toronto, sees customer engagement drop in winter but spike during Black Friday. She explores using purchase history and seasonal patterns to predict which products will perform best." \\
\midrule
\rowcolor{gray!3}
\textbf{Psychology} & "Raj notices truck drivers get stressed during monsoons, causing more accidents near the steel plant. He wonders how understanding driver behavior could create a safer work environment." & "Emma studies why customers make purchasing decisions differently based on mood and time of day. She explores how emotional triggers influence buying behavior for better campaigns." \\
\bottomrule
\end{tabular}
\begin{tablenotes}
\footnotesize
\item Examples follow narrative structure with character background, specific location context, problem identification, and solution exploration. Raj (Graduate, Industrial context), Emma (Post-Graduate, Corporate context).
\end{tablenotes}
\end{threeparttable}
\end{table}

\begin{table}[t]
\centering
\footnotesize
\begin{threeparttable}
\caption{Example Variations Based on Education Level}
\label{tab:education}
\begin{tabular}{>{\raggedright\columncolor{gray!8}}p{0.18\columnwidth} >{\raggedright\arraybackslash}p{0.32\columnwidth} >{\raggedright\arraybackslash}p{0.32\columnwidth}}
\toprule
\rowcolor{gray!25}
\textbf{Topic} & \textbf{Arjun (High School Student)} & \textbf{Kavya (Graduate Student)} \\
\midrule
\rowcolor{green!3}
\textbf{Machine Learning} & "Arjun, a high school student in Mumbai, notices Spotify suggests songs he likes. When he skips rock but replays pop songs, it shows more pop recommendations. He wonders how the app learns his taste." & "Kavya, a graduate student in Chennai, predicts real estate prices for her thesis. She collects housing data and trains neural networks using TensorFlow, experimenting with different architectures to minimize prediction errors." \\
\bottomrule
\end{tabular}
\begin{tablenotes}
\footnotesize
\item Examples follow narrative structure adapted to education level. Arjun (High School, Mumbai, relatable scenarios) uses everyday language and familiar contexts, while Kavya (Graduate, Chennai, academic context) includes technical terminology and research methodology.
\end{tablenotes}
\end{threeparttable}
\end{table}

Tables~\ref{tab:profession} and~\ref{tab:education} demonstrate how static personalization creates contextually appropriate examples while maintaining pedagogical objectives. Table~\ref{tab:profession} shows professional vocabulary adaptation, while Table~\ref{tab:education} illustrates complexity scaling based on educational background. This hybrid approach balances individual identity with real-time learning state, ensuring examples remain culturally and professionally relevant while adapting to current comprehension levels.

\section{Live Demonstration Walkthrough}

The demonstration will showcase ExaCraft through three interactive scenarios that highlight key behavioral adaptation features. Attendees will observe the system in action through a browser session with pre-configured user profiles and learning contexts, experiencing the multi-flow architecture illustrated in Figure \ref{fig:system-architecture}.

\subsection{Scenario 1: Adaptive Complexity Progression}

\textbf{Demo Setup:} The demonstration will begin with a simulated user (Profile: Graduate Student, Computer Science) browsing a machine learning tutorial webpage. Attendees will observe how personalized examples evolve through the following interaction sequence:

\begin{enumerate}
    \item User highlights machine learning → ExaCraft generates an intermediate-level personalized example leveraging the user's Computer Science academic background with programming analogies, algorithmic concepts, and cultural relevance (blue flow)
    \item User quickly progresses to neural networks and deep learning → ExaCraft detects mastery signals and adapts future examples with more sophisticated technical illustrations and implementation details
    \item User returns to machine learning and clicks the regenerate button twice → ExaCraft records struggle signals (orange flow) and generates simplified personalized examples with foundational analogies and step-by-step conceptual breakdowns
    \item ExaCraft demonstrates how personalized examples transform from technical implementations to conceptual illustrations based on behavioral learning patterns captured through the system architecture
\end{enumerate}

\textbf{Key Observable Features:} Attendees will observe how personalized examples change in complexity based on behavioral signals, demonstrating how dynamic adaptation works alongside static profiles to adjust example difficulty while maintaining cultural and professional relevance.

\subsection{Scenario 2: Session-Aware Learning Continuity}

\textbf{Demo Setup:} The demonstration will showcase cross-session personalized example continuity through a multi-part scenario:

\begin{enumerate}
    \item User initiates a focused learning session on economics → ExaCraft generates personalized examples using marketing contexts familiar to the user (green flow for session management)
    \item ExaCraft maintains personalized example preferences and complexity levels across topic progression through persistent context storage
    \item Session is ended and a new session is started after 3 minutes → ExaCraft recalls previous example topics and adapts based on established topic familiarity patterns (demonstrating cross-session continuity)
    \item User encounters difficulty with the topic supply and demand → ExaCraft recognizes previous familiarity with market research topic and generates examples that leverage familiar consumer behavior concepts to introduce pricing mechanisms, adapting complexity based on the established learning pattern
\end{enumerate}

\textbf{Key Observable Features:} The demonstration shows how personalized example preferences and topic familiarity patterns are maintained across sessions, enabling consistent yet adaptive personalized learning experiences that build upon previously explored concepts.

\subsection{Scenario 3: Multi-Signal Struggle Detection}

\textbf{Demo Setup:} The final scenario demonstrates sophisticated struggle detection for personalized example adaptation through multiple behavioral signals:

\begin{enumerate}
    \item User explores quantum computing with normal interaction patterns → ExaCraft generates standard complexity personalized examples appropriate to their profile (blue flow)
    \item User exhibits struggle signals: multiple regenerations, topic revisitation → ExaCraft automatically adapts to provide simplified personalized examples with familiar analogies and foundational conceptual illustrations (orange flow activation)
    \item Console log output demonstrates how multiple behavioral signals combine to trigger personalized example simplification while maintaining individual learning context.
\end{enumerate}

\textbf{Key Observable Features:} Demonstration of behavioral signal detection for personalized example adaptation, showing how struggle signals trigger simplified examples while maintaining individual learning preferences.

\section{Implementation and Deployment}

ExaCraft utilizes modern web technologies, ensuring broad compatibility and ease of deployment. The Chrome extension implements Manifest V3, while the Python Flask API utilizes Google Gemini AI via the LangChain framework. Privacy and data storage follow a local-first design: user profiles are kept in Chrome’s local storage, while learning context data (struggle/mastery indicators) is stored server-side in JSON format. Only minimal data like target topics and user profiles is transmitted for example generation. The demonstration provides attendees with direct ExaCraft interaction through the live browser extension, enabling firsthand experience of how dynamic learning context adaptation enhances traditional static personalization approaches.

\section{Conclusion}

ExaCraft addresses the problem of static, one-size-fits-all examples in educational content. Its core innovation is a dual-layer personalization framework that combines a learner's static profile (e.g., culture, profession) with dynamic behavioral analytics. By adapting to real-time interactions, ExaCraft provides examples that are more contextually relevant, enabling learners to relate to them more effectively than to generic alternatives.  
Our demonstration shows how ExaCraft adapts example complexity in real time and maintains user profiles across sessions using JSON-based storage for continuous personalization.

\section{Demo Access Information}

Video Demo: \url{https://youtu.be/jWy6uN9DoZw}\\
Live Demo Requirements: Standard laptop with Chrome browser, internet connection\\
Source Code: \url{https://github.com/akaash897/ExaCraft_Personalized_Example_Generation}

\begin{acks}
The authors would like to acknowledge grant no. 4(2)/2024-ITEA of MeitY, GoI; and Srijan: Center for Generative AI (grant no. ET/23/2024-ET) of MeitY under the IndiaAI mission with the support of Meta for partial support. Akaash Chatterjee gratefully acknowledges the IndiaAI Fellowship (D.O.~No.~INDAI/5/2025-INDAI) and the continued support of MeitY and the IndiaAI Mission.
\end{acks}

\bibliographystyle{ACM-Reference-Format}
\bibliography{sample-base.bib}

\end{document}